\documentclass[letterpaper]{article} 
\usepackage{aaai25}  
\usepackage{times}  
\usepackage{helvet}  
\usepackage{courier}  
\usepackage[hyphens]{url}  
\usepackage{graphicx} 
\usepackage{natbib}  
\usepackage{caption} 
\usepackage{algorithm}
\usepackage{algorithmic}

\usepackage{newfloat}
\usepackage{listings}

\usepackage{amsmath}
\usepackage{amsfonts}
\usepackage{multirow}
\usepackage{booktabs}

\DeclareCaptionStyle{ruled}{labelfont=normalfont,labelsep=colon,strut=off} 
\floatstyle{ruled}
\newfloat{listing}{tb}{lst}{}
\floatname{listing}{Listing}
\title{FSTA-SNN:Frequency-Based Spatial-Temporal Attention Module for Spiking Neural Networks}
\author{
    Kairong Yu\textsuperscript{\rm 1}\equalcontrib,
    Tianqing Zhang\textsuperscript{\rm 2}\equalcontrib,
    Hongwei Wang\textsuperscript{\rm 1}\equalcorresponding,
    Qi Xu\textsuperscript{\rm 3}\equalcorresponding
}
\affiliations{
    \textsuperscript{\rm 1}Zhejiang University-University of Illinois Urbana Champaign Institute, Zhejiang University, Haining, China \\
    \textsuperscript{\rm 2}The College of Computer Science and Technology, Zhejiang University, Hangzhou, China \\
    \textsuperscript{\rm 3}School of Computer Science and Technology, Dalian University of Technology, Dalian, China

    kairong.22@intl.zju.edu.cn, zhangtianqing@zju.edu.cn, hongweiwang@intl.zju.edu.cn, xuqi@dlut.edu.cn
}

\usepackage{bibentry}

\begin{document}
\maketitle

\begin{abstract}
Spiking Neural Networks (SNNs) are emerging as a promising alternative to Artificial Neural Networks (ANNs) due to their inherent energy efficiency.
Owing to the inherent sparsity in spike generation within SNNs, the in-depth analysis and optimization of intermediate output spikes are often neglected.
This oversight significantly restricts the inherent energy efficiency of SNNs and diminishes their advantages in spatiotemporal feature extraction, resulting in a lack of accuracy and unnecessary energy expenditure.
In this work, we analyze the inherent spiking characteristics of SNNs from both temporal and spatial perspectives.
In terms of spatial analysis, we find that shallow layers tend to focus on learning vertical variations, while deeper layers gradually learn horizontal variations of features.
Regarding temporal analysis, we observe that there is not a significant difference in feature learning across different time steps.
This suggests that increasing the time steps has limited effect on feature learning.
Based on the insights derived from these analyses, we propose a \textbf{F}requency-based \textbf{S}patial-\textbf{T}emporal \textbf{A}ttention (FSTA) module to enhance feature learning in SNNs.
This module aims to improve the feature learning capabilities by suppressing redundant spike features.
The experimental results indicate that the introduction of the FSTA module significantly reduces the spike firing rate of SNNs, demonstrating superior performance compared to state-of-the-art baselines across multiple datasets.
\end{abstract}

%
\begin{links}
    \link{Code}{https://github.com/yukairong/FSTA-SNN}
\end{links}

\section{Introduction}
Deep learning has achieved significant advancements in tasks like classification \cite{krizhevsky2012imagenet}, object segmentation \cite{ronneberger2015u}, and natural language processing \cite{hinton2012deep}, leading to state-of-the-art performance.
However, the high energy consumption of these architectures challenges hardware implementation. 
Spiking Neural Networks (SNNs) offer a promising alternative, inspired by the mammalian brain’s learning mechanisms.
SNNs represent a shift in information encoding and feature transmission by simulating the spatio-temporal dynamics of biological neurons \cite{roy2019towards, schuman2022opportunities}.
In SNNs, neurons theoretically fire only when the membrane potential surpasses a threshold, enabling event-driven operations on neuromorphic chips \cite{pei2019towards, ma2017darwin}.
This spike-based binary transmission allows SNNs to perform low-cost synaptic accumulations and avoid computations involving zero inputs or activations \cite{eshraghian2023training, deng2020rethinking}.

\begin{figure}[t]
    \centering
    \includegraphics[width=1.0\linewidth]{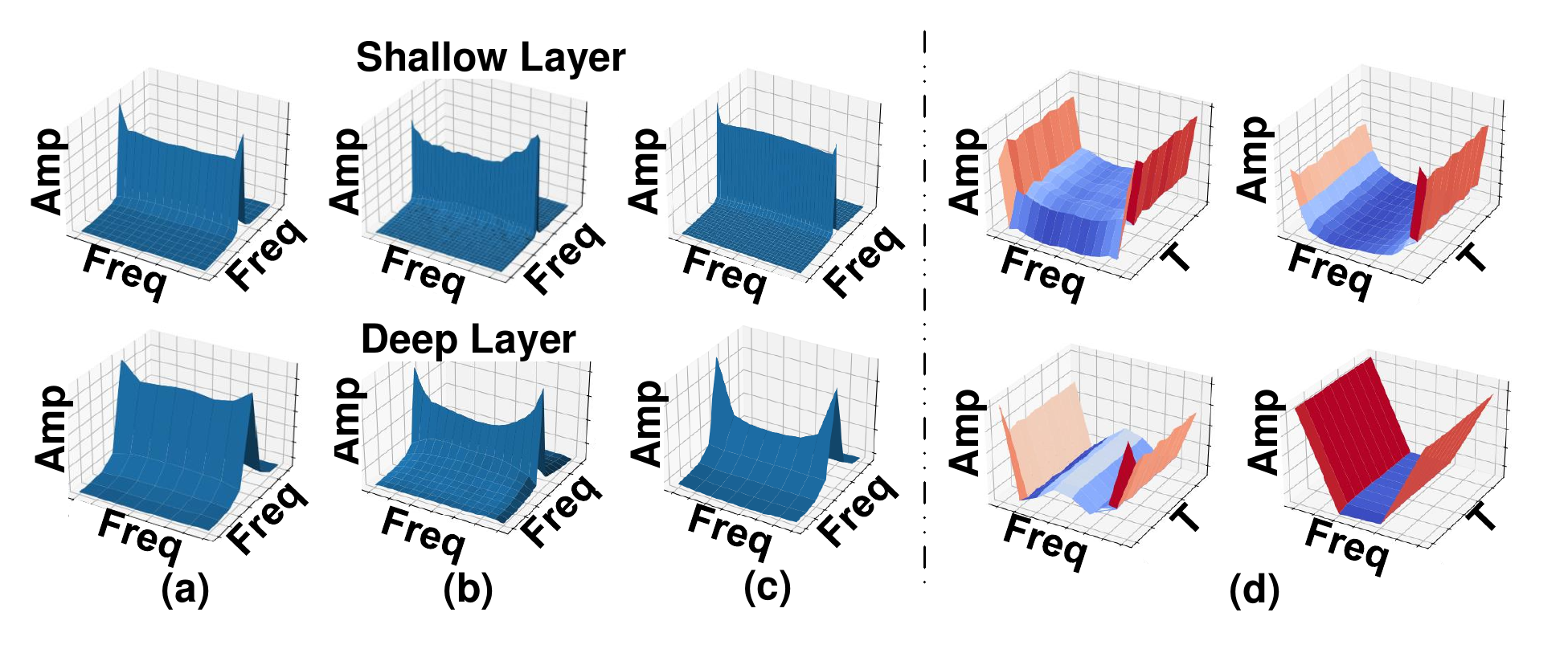}
    \caption{Comparison of the output spike frequency spectrum distribution of SNNs across different model structures, datasets, layer depths, and temporal perspectives.
    (a) The spectral distribution of spike outputs at different layers of the ResNet architecture at the same time step in static dataset.
    (b) The spectral distribution of spike outputs at different layers of the ResNet architecture at the same time step in dynamic dataset.
    (c) The spectral distribution of spike outputs at different layers of the VGG architecture at the same time step in static dataset.
    (d) The temporal variation of the frequency bands where spectral energy is most concentrated across different layers.
    Here, Amp, Freq, and T represent amplitude, frequency range and timesteps.
    }
    \label{fig:activation}
\end{figure}
The common belief that computations in SNNs, which are pulse-based neuromorphic computing structures, are inherently sparse is disputable.
Due to the lack of reliable mathematical theoretical tools, explorations and analyses in this area are limited.
Current investigations predominantly rely on lightweight, spike counting or attention-based strategies.
For example, proposed algorithms incorporate penalty functions to exploit spike-aware sparse regularization and compression \cite{2021Comprehensive, yin2021accurate},
employ techniques like pruning and distillation to reduce spike occurrences \cite{shen2023esl,xu2023constructing,xu2023biologically}
or integrate attention mechanisms like temporal, spatial, or channel attention to suppress redundant spikes \cite{yao2023inherent, yao2023attention}.
Despite the potential of these approaches to decrease spikes, they are associated with drawbacks such as decreased accuracy, limited enhancements, or the introduction of additional complexity, without conducting a comprehensive network-level assessment of learning preferences.

In this work, we analyze the learning preferences of spike discharges in SNNs from a frequency perspective, providing a novel understanding of spike redundancy.
As depicted in Fig. \ref{fig:activation}, to validate the universality of learning in SNNs, we conduct Fourier transforms on spike outputs from intermediate layers of varying depths across different datasets and model architectures.
The spectral analysis reveal consistent frequency distributions in both ResNet and VGG structures across static and dynamic datasets.
Over time, the frequency distributions across the same layer exhibits significant overlap with only minor variations in amplitude response, suggesting potential parameter sharing to enhance efficiency spatially.
Observing network depth, shallow layers are concentrated along a central axis, while deeper layers gradually spread along a vertical axis.
This phenomenon suggests that in shallow layers, SNNs preferentially discharge spikes for vertical disparity features, gradually shifting towards horizontal disparity features in deeper layers.

Based on these findings, we propose a novel frequency-based spatial-temporal attention (FSTA) module. 
This module enhances the learning capabilities of existing networks by amplifying previously overlooked spike features while preserving the network’s inherent fitting preferences. 
Additionally, it effectively suppresses redundant spike features and irrelevant noise. 
Experimental results show that our module significantly improves SNN performance, reduces spike firing rate, and maintains computational efficiency without increasing energy consumption.

Overall, our main contributions are threefold:
\begin{itemize}
    \item We present a comprehensive study of SNN learning preferences, introducing a novel frequency-based spike analysis.
    This framework is essential for optimizing sparsification and energy efficiency and provides a theoretical foundation for enhancing SNN performance. 

    \item We propose a Frequency-Based Spatial-Temporal Attention (FSTA) module, a plug-and-play component that introduces a minimal number of additional parameters.
    This module effectively reduces spike firing rate while boosting performance.
    
    \item We evaluate our method on static datasets CIFAR10, CIFAR100, ImageNet, and dynamic dataset CIFAR10-DVS using widely adopted architectures.
    The results show that networks integrating our proposed module achieve state-of-the-art performance, with a total spike firing rate reduction of approximately 33.99\%.
    
\end{itemize}

\section{Related Works}
\paragraph{Frequency Domain Learning in ANNs}
The effectiveness of frequency applications in deep learning tasks is well-established, as real-world datasets often contain rich frequency information.
In the ANN domain, approaches that integrate frequency into network architectures can be broadly categorized into two types.
One approach involves transforming the model from the spatial domain to the frequency domain for direct learning 
 \cite{huang2023adaptive, patro2023spectformer, guo2023spatial, kong2023efficient}.
These methods typically start with applying Fourier 
 \cite{brigham1988fast} or Wavelet Transforms \cite{burrus1998wavelets} to the original feature maps, followed by linear mappings to convert frequency information, and end with an inverse transform to restore the features.
The other approach leverages frequency information for local feature extraction within the network.
For example, in \cite{qin2021fcanet}, the introduction of Discrete Cosine Transform (DCT) \cite{ahmed1974discrete} into the channel attention mechanism SENet \cite{hu2018squeeze} enhances the channel's ability to perceive more comprehensive frequency information.

\paragraph{Frequency Applications in SNNs}
The unique advantages of frequency have led to various application methods in SNNs.
Current approaches include using frequency for temporal encoding, such as employing DCT to minimize the number of time steps required for inference \cite{garg2021dct} and introducing spike frequency-adaptive neurons to dynamically adjust neuronal firing thresholds \cite{chen2023improving, falez2018mastering}.
In contrast to these previous methods, our focus is on leveraging frequency-based activation of network output spikes.
Furthermore, we propose the integration of a frequency-based attention module between layers to minimize redundancy and reduce the spike firing rate.

\paragraph{Attention in SNNs}
Attention mechanisms in deep learning have achieved significant success, primarily motivated by the human ability to efficiently focus on salient information within complex scenarios.
A common approach is to incorporate attention as an auxiliary module to enhance the representational capacity of ANNs \cite{li2022ham,guo2022attention}.
This work \cite{yao2021temporal} involves using temporal attention to evaluate the importance of different time steps while bypassing non-essential ones.
Furthermore, current research is increasingly exploring multi-dimensional attention modules that encompass time, space, and channels \cite{zhu2024tcja,yao2023attention,xu2024enhancing}.
Although these methods effectively reduce spike firing and improve accuracy, they also introduce substantial additional computational burdens to SNNs.
\begin{figure*}[t]
    \centering
    \includegraphics[width=1.0\linewidth]{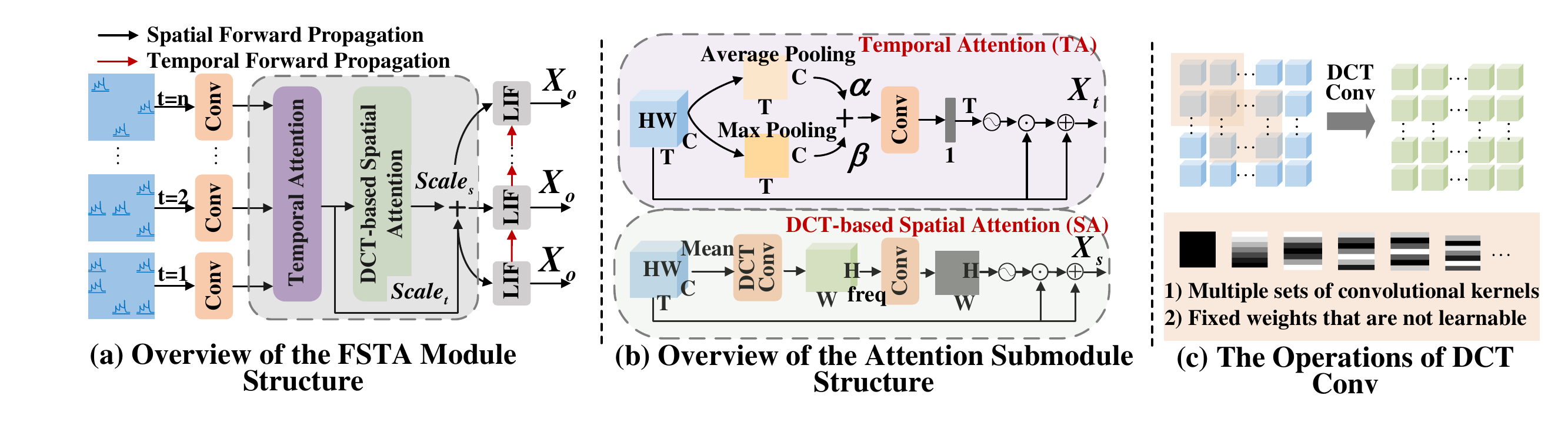}
    \caption{Overview of the FSTA module and its internal submodules structure}
    \label{fig:framework}
\end{figure*}

\section{SNN Frequency Analysis}
\paragraph{SNN Fundamentals}
The fundamental computational unit of SNNs is the spiking neuron, 
an abstract representation of biological neuronal dynamics.
The Leaky Integrate-and-Fire (LIF) model is among the most widely used spiking neuron models, as it effectively balances simplified mathematical representations with the complex dynamics of biological neurons.
Mathematically, an LIF neuron is described by the following equation:
\begin{equation}
     \tau \frac{du(t)}{dt} = -(u(t) - u_\text{reset}) + I(t)
\label{eq:1}
\end{equation}
where \(\tau\) denotes the membrane potential time constant, \(u(t)\) represents the membrane potential at time step \(t\), \(u_\text{reset}\) is the neuron's resting potential, and \(I(t)\) is the synaptic input at time step \(t\).
According to Eq.\ref{eq:1}, the discrete-time and iterative mathematical representation of LIF-SNNs can be described as follows:
\begin{align}
    & V^{t,n} = H^{t-1, n} + \frac{1}{\tau} [I^{t-1,n} - (H^{t-1, n} - V_{reset})] \\
    & S^{t,n} = \Theta(V^{t,n} - v_{th}) \\
    & H^{t,n} = V_{reset} \cdot S^{t,n} + V^{t,n} \odot (1 - S^{t,n}). 
\end{align}
The Heaviside step function \(\Theta\) is defined as:
\begin{equation}
    \Theta(x) = 
        \begin{cases}
            0 & \text{if } x < 0 \\
            1 & \text{if } x \geq 0.
        \end{cases}
\end{equation}
Among these, \(H^{t-1,n}\) denotes the membrane potential after a spike from the previous time step. \(I^{t,n}\) and \(V^{t,n}\) represent the input and updated membrane potential at time step \(t\) for \(n\)-th layer, respectively.
Additionally, \(v_{th}\) is the threshold that determines whether \(V^{t,n}\) results in a spike or remains silent, and \(S^{t,n}\) indicates the spike sequence at time step \(t\) for \(n\)-th layer.

\paragraph{Frequency Fundamentals}
Frequency analysis methods are widely used for feature analysis, with the Discrete Fourier Transform (DFT) being one of the most popular algorithms.
The DFT is pivotal in digital signal processing and serves as an essential analytical tool for the preference learning of SNNs discussed in this paper.
For clarity, we first consider the one-dimensional DFT.
Given a sequence of \(N\) complex numbers \(x[n]\), where
\(n = 0, 1, \dots,N-1\),
the one-dimensional DFT converts this sequence into the frequency domain as follows:
\begin{equation}
\centering
    X[k] = \sum^{N-1}_{n=0}x[n] e^{-j(2 \pi / N) k n} 
\label{eq:dft-1d}
\end{equation}
where \(j\) is the imaginary unit. Eq. \ref{eq:dft-1d} can be derived by performing the Fourier transform of a discrete signal through sampling in both the time domain and the frequency domain.
Since \(X[k]\) repeats with a period of length \(N\), it is sufficient to evaluate \(X[k]\) at \(N\) discrete points  at \(k=0,1,...,N-1\).
Specifically, \(X[k]\) represents the frequency spectrum of the sequence \(x[n]\) at the frequency \(\omega_k = \frac{2 \pi k}{N}\).

It is also noteworthy that the DFT is a one-to-one transformation.
Given \(X[k]\), the original signal \(x[n]\) can be reconstructed using the Inverse Discrete Fourier Transform (IDFT).
\begin{equation}
\centering
    x[n] = \frac{1}{N} \sum^{N-1}_{k=0} X[k] e^{j (2 \pi / N ) k n}
\label{eq:idft-1d}
\end{equation}
The DFT can similarly be extended to 2D signals.
Given a 2D signal \(X[m,n]\), where \(0 \leq m \leq M-1\) and \(0 \leq n \leq N-1\), the 2D DFT of \(x[m,n]\) is given by the following formula:
\begin{equation}
    X[u,v] = \sum^{M-1}_{m=0} \sum^{N-1}_{n=0} x[m,n] e^{-j 2 \pi (\frac{um}{M} + \frac{vn}{N})}
\end{equation}
The 2D DFT can be considered as the sequential application of 1D DFTs in two dimensions.

\paragraph{Frequency Analysis}
Building on the foundational knowledge outlined above, we analyze the SNN. 
To accurately represent the network's learning preferences, we perform a statistical analysis of spike outputs from various intermediate layers across different datasets.
We use the DFT for frequency domain conversion and center the results, as shown in Fig. \ref{fig:activation}.
Based on these frequency spectra, we investigate the frequency characteristics of the SNNs across spatial, temporal, and other dimensions.

\paragraph{Observation 1. The structural differences and dataset variations do not affect the learning preferences of SNNs}
\label{sec:obs1}

As shown in Fig. \ref{fig:activation}(a), (b), and (c), the frequency distributions of spike average release probabilities are remarkably similar across different architectures and datasets.
In shallow layers, the spectral bands of SNNs are highly concentrated along the central horizontal axis, capturing nearly all spectral energy.
In deeper layers, network frequency gradually shifts toward the vertical axis, although the central horizontal axis remains significant.
According to Eq.\ref{eq:idft-1d}, the spectrum suggests that SNNs initially focus on longitudinal differential features in shallow layers and progressively shift to capturing lateral differential features in deeper layers.

\paragraph{Observation 2: Increasing the time step has limited potential to enhance performance}
\label{sec:obs2}
From Fig. \ref{fig:activation}(d), it is evident that the frequency distribution remains consistent across different time steps, with only minor variations in amplitude.
This consistency becomes more pronounced in deeper layers, indicating that features learned within the same layer stabilize over time and suggesting limited acquisition of new feature information beyond a certain time step.
Moreover, this observation suggests that spatial enhancement module across different time steps could share parameters, thereby reducing redundancy.

\section{Methodology}
Based on the previous analysis and conclusions, we majorly focus on four issues as follows:
(1) Sharing enhancement modules across identical frequency distributions at different time steps within the same layer.
(2) Reducing redundancy in feature learning within shallow networks.
(3) Expanding feature learning capacity in deep networks.
(4) Adjusting amplitude variations across time steps using temporal attention.
To address these issues, we employ a Frequency-based Spatial-Temporal Attention (FSTA) module.

\begin{table*}[!h]
\centering
\small
\begin{tabular}{cccccc}
\toprule
Dataset                     & Method                         & Type                                   & Architecture                       & Timestep   & Accuracy         \\ \hline
\multirow{26}{*}{CIFAR-10}  & SpikeNorm \cite{sengupta2019going}                     & ANN2SNN                                & VGG16                              & 2500       & 91.55\%          \\
                            & Hybrid-Train \cite{rathi2020enabling}                  & Hybrid training                        & VGG16                              & 200        & 92.02\%          \\
                            & PTL \cite{wu2021progressive}                           & Tandem learning                        & VGG11                              & 16         & 91.24\%          \\
                            & DSR \cite{meng2022training}                          & SNN training                           & ResNet18                           & 20         & 95.40\%          \\
                            & KDSNN \cite{xu2023constructing}                        & SNN training                           & ResNet18                           & 4          & 93.41\%          \\
                            & Joint A-SNN \cite{guo2023joint}                   & SNN training                           & ResNet18                           & 4          & 95.45\%          \\ \cline{2-6} 
                            & \multirow{3}{*}{RMP-Loss \cite{guo2023joint}}      & \multirow{3}{*}{SNN training}          & \multirow{2}{*}{ResNet19}          & 2          & 95.31\%          \\
                            &                                &                                        &                                    & 4          & 95.51\%          \\ \cline{4-6} 
                            &                                &                                        & ResNet20                           & 4          & 91.89\%          \\ \cline{2-6} 
                            & \multirow{2}{*}{RecDis-SNN \cite{guo2022recdis}}    & \multirow{2}{*}{SNN training}          & \multirow{2}{*}{ResNet19}          & 2          & 93.64\%          \\
                            &                                &                                        &                                    & 4          & 95.53\%          \\ \cline{2-6} 
                            & \multirow{2}{*}{TET \cite{deng2022temporal}}           & \multirow{2}{*}{SNN training}          & \multirow{2}{*}{ResNet19}          & 2          & 94.16\%          \\
                            &                                &                                        &                                    & 4          & 94.44\%          \\ \cline{2-6} 
                            & \multirow{3}{*}{Real Spike \cite{guo2022real}}    & \multirow{3}{*}{SNN training}          & \multirow{3}{*}{ResNet19}          & 2          & 95.31\%          \\
                            &                                &                                        &                                    & 4          & 95.51\%          \\
                            &                                &                                        &                                    & 6          & 96.10\%          \\ \cline{2-6} 
                            & \multirow{5}{*}{MPBN \cite{guo2023membrane}}          & \multirow{5}{*}{SNN training}          & \multirow{2}{*}{ResNet19}          & 1          & 96.06\%          \\
                            &                                &                                        &                                    & 2          & 96.47\%          \\ \cline{4-6} 
                            &                                &                                        & \multirow{3}{*}{ResNet20}          & 1          & 92.22\%          \\
                            &                                &                                        &                                    & 2          & 93.54\%          \\
                            &                                &                                        &                                    & 4          & 94.28\%          \\ \cline{2-6} 
                            & \multirow{5}{*}{\textbf{Ours}} & \multirow{5}{*}{\textbf{SNN training}} & \multirow{2}{*}{\textbf{ResNet19}} & \textbf{1} &       $\textbf{96.21\%} \pm 0.10\%$ \\
                            &                                &                                        &                                    & \textbf{2} &  $\textbf{96.52\%} \pm 0.09\%$ \\ \cline{4-6} 
                            &                                &                                        & \multirow{3}{*}{\textbf{ResNet20}} & \textbf{1} & $\textbf{93.01\%} \pm 0.12\%$ \\
                            &                                &                                        &                                    & \textbf{2} & $\textbf{94.18\%} \pm 0.10\%$ \\
                            &                                &                                        &                                    & \textbf{4} & $\textbf{94.72\%} \pm 0.09\%$ \\ \hline
\multirow{21}{*}{CIFAR-100} & RMP \cite{han2020rmp}                           & ANN2SNN                                & ResNet20                           & 2048       & 67.82\%          \\
                            & LTL  \cite{yang2022training}                          & Tandem learning                        & ResNet20                           & 31         & 76.08\%          \\ 
                            & Real Spike \cite{guo2022real}                   & SNN training                           & ResNet20                           & 5          & 70.62\%          \\ \cline{2-6}
                            & \multirow{2}{*}{Dspike \cite{li2021differentiable}}        & \multirow{2}{*}{SNN training}          & \multirow{2}{*}{ResNet20}          & 2          & 71.68\%          \\
                            &                                &                                        &                                    & 4          & 73.35\%          \\ \cline{2-6} 
                            & \multirow{2}{*}{TET \cite{deng2022temporal}}           & \multirow{2}{*}{SNN training}          & \multirow{2}{*}{ResNet19}          & 2          & 72.87\%          \\
                            &                                &                                        &                                    & 4          & 74.47\%          \\ \cline{2-6} 
                            & \multirow{2}{*}{GLIF \cite{yao2022glif}}          & \multirow{2}{*}{SNN training}          & \multirow{2}{*}{ResNet19}          & 2          & 75.48\%          \\
                            &                                &                                        &                                    & 4          & 77.05\%          \\ \cline{2-6} 
                            & \multirow{3}{*}{TEBN \cite{duan2022temporal}}          & \multirow{3}{*}{SNN training}          & \multirow{3}{*}{ResNet19}          & 2          & 75.86\%          \\
                            &                                &                                        &                                    & 4          & 76.13\%          \\
                            &                                &                                        &                                    & 6          & 76.41\%          \\ \cline{2-6}  
                            & \multirow{4}{*}{MPBN \cite{guo2023membrane}}          & \multirow{4}{*}{SNN training}          & \multirow{2}{*}{ResNet19}          & 1          & 78.71\%          \\
                            &                                &                                        &                                    & 2          & 79.51\%          \\ \cline{4-6} 
                            &                                &                                        & \multirow{2}{*}{ResNet20}          & 2          & 70.79\%          \\
                            &                                &                                        &                                    & 4          & 72.30\%          \\ \cline{2-6} 
                            & \multirow{5}{*}{\textbf{Ours}} & \multirow{5}{*}{\textbf{SNN training}} & \multirow{2}{*}{\textbf{ResNet19}} & \textbf{1} & $\textbf{78.87\%} \pm 0.11\%$ \\
                            &                                &                                        &                                    & \textbf{2} & $\textbf{80.42\%} \pm 0.09\%$ \\ \cline{4-6} 
                            &                                &                                        & \multirow{3}{*}{\textbf{ResNet20}} & \textbf{1} & $\textbf{69.64\%} \pm 0.10\%$ \\
                            &                                &                                        &                                    & \textbf{2} & $\textbf{72.15\%} \pm 0.12\%$ \\
                            &                                &                                        &                                    & \textbf{4} & $\textbf{73.44\%} \pm 0.09\%$ \\ 
\bottomrule
\end{tabular}
\caption{Comparison with SOTA methods on CIFAR-10/100}
\label{tab:cifar10/100}
\end{table*}

\subsection{DCT-based Spatial Attention Submodule}
We begin with a detailed explanation of the Discrete Cosine Transform (DCT) \cite{ahmed1974discrete}.
The basis function of the two-dimensional (2D) DCT is defined as:
\begin{equation}
    B^{i,j}_{u,v} = \cos{(\frac{\pi u}{H} (i + \frac{1}{2}))} \cos{(\frac{\pi v}{W} (j + \frac{1}{2}))}
\label{eq:dct-basis}
\end{equation}
The 2D DCT can then be expressed as:
\begin{equation}
    f_{u,v} = \sum^{H-1}_{i=0} \sum^{W-1}_{j=0} x_{i,j} B^{i,j}_{u,v} 
\label{eq:dct-2d}
\end{equation}
where \( u \in \{0,1,..., H-1\}, v \in \{0,1,...,W-1\} \).
Here, \(f \in \mathbb{R}^{H,W}\) represents the 2D DCT frequency spectrum, while \(x \in \mathbb{R}^{H,W}\) denotes the input, with \(H\) and \(W\) being the height and width of \(x\), respectively.

In typical attention modules, global average pooling (GAP) is widely used for its computational simplicity and effective compression. 
Other methods such as global max pooling and global standard pooling are also common. 
However, as shown in Eq. \ref{eq:dct-2d}, the addition of \(x\) and \(B\) closely resembles convolution operations.
This establishes a link between traditional feature extraction methods and frequency domain analysis.
We will demonstrate that the compression method used in GAP is, in fact, a special case of 2D DCT, with results proportional to the lowest frequency component of the 2D DCT.

\textit{Proof.}
Assume \(u\) and \(v\) in Eq. \ref{eq:dct-2d} are both set to 0. Then:
\begin{align}
    f_{0,0} & = \sum^{H-1}_{i=0} \sum^{W-1}_{j=0} x_{i,j} \cos{(\frac{0}{H}(i+\frac{1}{2}))} \cos{(\frac{0}{W}(j+\frac{1}{2}))}  \notag \\
    & = \sum^{H-1}_{i=0} \sum^{W-1}_{j=0} x_{i,j} = GAP(x)HW 
\label{eq:gap-proof}
\end{align}

Since \(H\) and \(W\) are constants, linear operations in neural network can be transformed through mapping layers without altering the feature distribution.
Consequently, from Eq. \ref{eq:gap-proof}, it is evident that GAP corresponds to the lowest frequency component of the 2D DCT.
This suggests that the spatial attention features previously used are confined to specific frequency bands and fail to capture comprehensive information.
Based on these insights, we propose a DCT-based full-spectrum spatial attention module that better aligns with learning preferences of SNNs.

As illustrated in Fig. \ref{fig:framework}(b), the first step involves averaging the input \(X\) along the temporal dimension.
This is because the frequency feature distributions in Obs. 2 \ref{sec:obs2} are highly similar across different time steps.
Consequently, the shape of \(X\) changes from \(\mathbb{R}^{T,C,H,W}\) to \(\mathbb{R}^{C,H,W}\).
Next, the feature map \(X_{mean}\) is analyzed using the full-band frequency basis of the DCT to extract the complete frequency feature \(Freq\), where \(Freq \in \mathbb{R}^{freq,H,W}\) and $freq$ represents the number of frequency components.
This entire extract process can be achieved through concatenation:
\begin{align}
\centering
    X_{mean} & = mean(X) \\
    Freq & = Conv_{dct}(X_{mean})
\end{align}
Here, \(Conv_{dct}\) represents a non-trainable kernel that performs a convolution operation using fixed weights, as derived from Eq. \ref{eq:dct-basis}.
In the experimental section, we evaluate how different kernel size, which correspond to various frequency extraction ranges, affect overall network performance.

After extracting the frequency features \(Freq\),
the linear layer compresses these features and applies the Sigmoid function to obtain the spatial attention weight matrix \(freq_w \in \mathbb{R}^{H,W}\),
which encodes where to emphasize or suppress.
Finally, \(freq_w\) is dot-multiplied with input \(X\) and added, thereby enhancing the feature matrix.
The complete process of compression and enhancement is as follows:
\begin{align}
\centering
     freq_w & = Sigmoid(Linear(Freq)) \\
     X_s & = X + X \cdot freq_w
\end{align}

To further reduce computational complexity and avoid complex floating-point operations, the weights of  \(Conv_{dct}\) are precomputed and stored as fixed constants, eliminating the need for repetitive cosine calculations.
This module design adds only a minimal number of parameters while avoiding excessive computational overhead.
Additionally, it broadens the frequency range of the extracted information, reducing spike redundancy and enhancing feature representation.

\begin{table}[h]
\centering
\small
\setlength{\tabcolsep}{5pt}
\begin{tabular}{cccc}
\toprule
Method                         & Architecture       & T   & Accuracy         \\ \hline
STBP-tdBN \shortcite{zheng2021going}                      & ResNet34          & 6          & 63.72\%          \\
TET \shortcite{deng2022temporal}                           & ResNet34          & 6          & 64.79\%          \\
RecDis-SNN \shortcite{guo2022recdis}                    & ResNet34          & 6          & 67.33\%          \\
GLIF \shortcite{yao2022glif}                          & ResNet34          & 4          & 67.52\%          \\
IM-Loss \shortcite{guo2022loss}                       & ResNet18          & 6          & 67.43\%          \\ \hline
\multirow{2}{*}{Real Spike \shortcite{guo2022real}}    & ResNet18          & 4          & 63.68\%          \\
                               & ResNet34          & 4          & 67.69\%          \\ \hline
\multirow{2}{*}{RMP-Loss \shortcite{guo2023joint}}      & ResNet18          & 4          & 63.03\%          \\
                               & ResNet34          & 4          & 65.17\%          \\ \hline
\multirow{2}{*}{MPBN \shortcite{guo2023membrane}}          & ResNet18          & 4          & 63.14\%          \\
                               & ResNet34          & 4          & 64.71\%          \\ \hline
\multirow{2}{*}{SEW ResNet \shortcite{fang2021deep}}    & ResNet18          & 4          & 63.18\%          \\
                               & ResNet34          & 4          & 67.04\%          \\ \hline
\multirow{2}{*}{\textbf{Ours}} & \textbf{ResNet18} & \textbf{4} & $\textbf{68.21\%} \pm 0.20\%$ \\
                               & \textbf{ResNet34} & \textbf{4} & $\textbf{70.23\%} \pm 0.12\%$ \\ 
\bottomrule
\end{tabular}
\caption{Comparison with SNN training based SOTA methods on ImageNet}
\label{tab:imagenet}
\end{table}

\begin{table}[h]
\centering
\small
\setlength{\tabcolsep}{5pt}
\begin{tabular}{cccc}
\toprule
Method                         & Architecture                       & T    & Accuracy         \\ \hline
DSR \shortcite{meng2022training}                           & VGG11                              & 20          & 77.27\%          \\
GLIF \shortcite{yao2022glif}                          & 7B-wideNet                         & 16          & 78.10\%          \\
STBP-tdBN \shortcite{zheng2021going}                     & ResNet19                           & 10          & 67.80\%          \\
RecDis-SNN \shortcite{guo2022recdis}                     & ResNet19                           & 10          & 72.42\%          \\
Real Spike \shortcite{guo2022real}                    & ResNet19                           & 10          & 72.85\%          \\
TET \shortcite{deng2022temporal}                           & VGGSNN                             & 10          & 77.30\%          \\ \hline
\multirow{2}{*}{MPBN \shortcite{guo2023membrane}}          & ResNet19                           & 10          & 74.40\%          \\
                               & ResNet20                           & 10          & 78.70\%          \\ \hline
\multirow{2}{*}{\textbf{Ours}} & \multirow{2}{*}{\textbf{ResNet20}} & \textbf{10} & $\textbf{81.50\%} \pm 0.25\%$ \\
                               &                                    & \textbf{16} & $\textbf{82.70\%} \pm 0.10\%$ \\ \bottomrule
\end{tabular}
\caption{Comparison with SNN training based SOTA methods on CIFAR10-DVS}
\label{tab:cifar10-dvs}
\end{table}

\subsection{Temporal Attention Submodule for Amplitude Regulation}
In Obs. 2\ref{sec:obs2}, it is noted that SNN layers do not exhibit frequency range fitting biases temporally, but rather amplitude differences.
Therefore, attention must be focused on learning spike features over time.
As depicted in Fig. \ref{fig:framework}(b), for the input spike feature map \(X\), temporal channel features are initially aggregated using average and max pooling operations.
To effectively integrate these temporal features, we introduce two learnable parameters \(\alpha\) and \(\beta\), which balance global (max pooling) and local (average pooling) information.
Specifically, \(f_{avg}, f_{max} \in \mathbb{R}^{T, 1, 1}\).
Feature extraction \(M\) is performed using the following equation:
\begin{equation}
\centering
    M = \alpha \cdot f_{avg}(X) + \beta \cdot f_{max}(X)
\end{equation}
where \(M \in \mathbb{R}^{T,C}\).

After extraction, the mean is computed across the temporal dimension of \(M\).
Subsequently, linear layers followed by a Sigmoid function are used to obtain weights \(T_w \in \mathbb{R}^{T}\) for different time steps.
Finally, as with the spatial module, temporal enhancement is applied to the input.
\begin{align}
\centering
    & M_{mean} = mean(M) \\
    & T_w = Sigmoid(Linear(M_{mean})) \\ 
    & X_t = X + X \cdot T_w
\end{align}

\subsection{Frequency-based Spatial-Temporal Attention Module}
After detailing the temporal attention (TA) submodule and the DCT-based spatial attention (SA) submodule,
we proceed to integrate the entire module.
Our experiments (see the experimental section for details) show that concatenating the TA and SA achieves the optimal combination,
balancing computational complexity and performance effectively.
For the intermediate output \(X\) in the SNN,
we first apply TA to enhance temporal dynamics, followed by spatial frequency enhancement using SA.
To address significant loss of intermediate feature information, we introduce essential temporal and spatial scaling factors \(Scale_{t}\) and \(Scale_{s}\) to control this process and obtain the fused output.
The complete computational process is outlined as follows:
\begin{align}
    X_t & = TA(X) \\
    X_s & = SA(X_t) \\
    X_o & = Scale_t \cdot X_t + Scale_s \cdot X_s
\end{align}
in which, \(X_t, X_s, X_o\) maintain the same shape as \(X\).

\section{Experiments}
We perform comprehensive experiments to assess the proposed method and compare it with other recent SOTA methods on several widely used architectures.
These experiments utilize static datasets CIFAR-10, CIFAR-100 \cite{krizhevsky_cifar-10_2010}, and ImageNet \cite{deng_imagenet_2009}, as well as the dynamic dataset CIFAR10-DVS \cite{li_cifar10-dvs_2017} dataset.

\subsection{Ablation Study}
\subsubsection{Combination Mode of Submodules}
\begin{figure}
    \centering
    \includegraphics[width=0.85\linewidth]{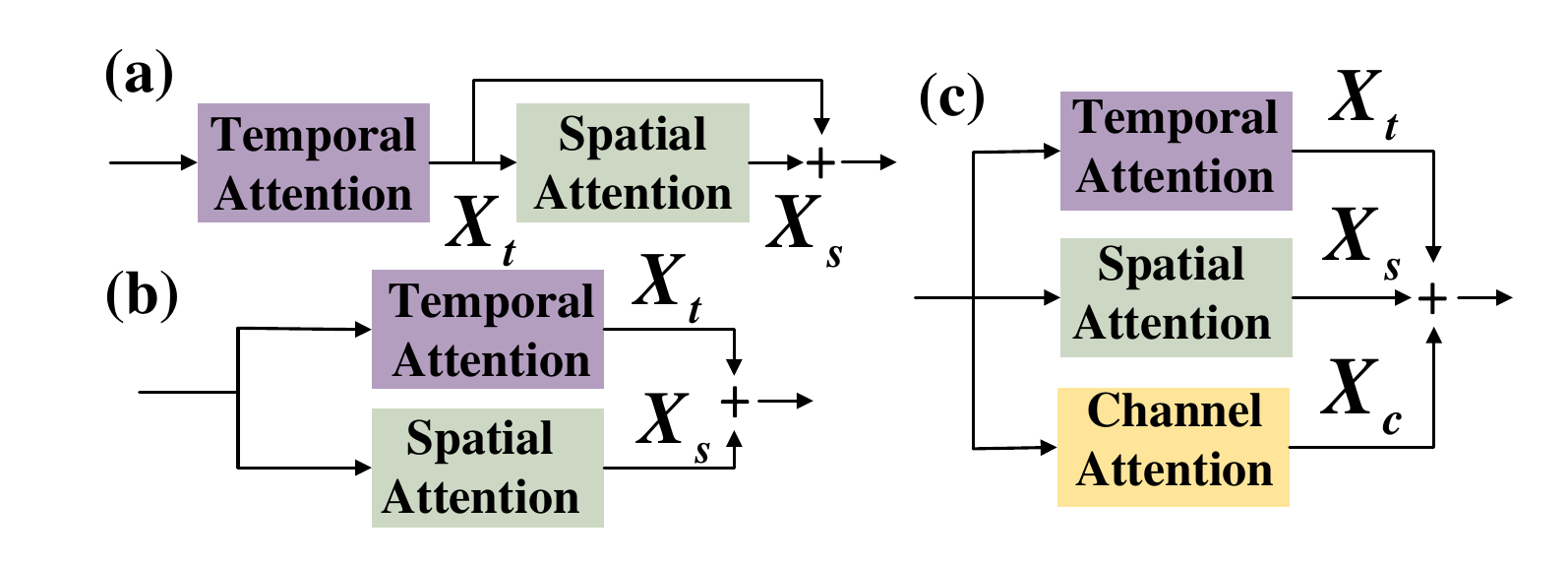}
    \caption{The different combinations of modules}
    \label{fig:modules-mode}
\end{figure}

To enhance module integration performance with the original SNN, we examine various submodule combination strategies.
In these experiments, we use ResNet20 as the baseline network and evaluate its performance on the CIFAR-100 dataset with a time step of 2.
As illustrated in Fig. \ref{fig:modules-mode}, we explore both parallel and serial combinations, and include a channel attention enhancement submodule for comparison.
Results in Tab. \ref{tab:modules-mode} show that the serial combination mode slightly improves accuracy by \(1.32\%\) over the parallel mode.
Although channel attention enhances performance, it increases the number of MAC operations and parameters, thereby raising energy consumption.
Ultimately, the serial combination of spatiotemporal attention provides the best balance between performance and energy efficiency.

\begin{table}[t]
\centering
\small
\begin{tabular}{cccc}
\toprule
Mode & ACs     & MACs   & Accuracy \\ \hline
a    & 109.91M & 38.93M & 72.15\%  \\
b    & 107.31M & 38.93M & 70.83\%  \\
c    & 89.17M  & 41.74M & 72.68\%  \\ \bottomrule
\end{tabular}
\caption{Comparative analysis of the computational performance of different combination methods}
\label{tab:modules-mode}
\end{table}

\begin{table*}[h]
\centering
\small
\begin{tabular}{ccccccccc}
\toprule
Dataset     & Architecture & Resolution & Timestep & ACs     & MACs    & FLOPs   & Param(M) & Energy(mJ) \\ \hline
CIFAR100    & ResNet20     & 32x32      & 4        & 260.05M & 67.26M  & 880.36M & 11.3     & 0.54   \\
CIFAR10-DVS & ResNet20     & 128x128    & 10       & 2.14G   & 582.87M & 8.71G   & 11.2     & 4.60   \\
ImageNet    & SEW ResNet18     & 224x224    & 4        & 2.45G   & 1.05G   & 7.37G   & 11.7     & 7.03   \\ \bottomrule
\end{tabular}
\caption{Energy costs and model structure across different datasets}
\label{tab:energy-cost}
\end{table*}

\subsubsection{Selection of Frequency Range}
Eq. \ref{eq:dct-2d} demonstrates that frequency feature extraction can be performed using convolutional operations.
The size of the convolutional kernel directly influences the frequency range of \(u\) and \(v\).
To enhance module versatility and reduce manual computation, \(Conv_{dct}\) uses a block-wise approach with fixed kernel sizes for feature extraction.
Thus, exploring various kernel shapes, corresponding to different frequency ranges, is justified for improving network performance.
In Tab. \ref{tab:kernel}, we compare different kernel sizes and strides to maintain input-output consistency.
With experimental conditions kept identical except for kernel sizes of \(3\times3, 5\times5\), and \(7\times7\), the performance results are \(92.36\%, 92.58\%\), and \(93.01\%\), respectively.
These results suggest that finer frequency ranges lead to improved performance within a certain range.

\begin{table}[t]
\centering
\small
\begin{tabular}{cccc}
\toprule
Kernel Size & Frequency Range & Padding & Accuracy \\ 
\midrule
3x3         & 9               & 1       & 92.36\%  \\
5x5         & 25              & 2       & 92.58\%  \\
7x7         & 49              & 3       & 93.01\%  \\ 
\bottomrule
\end{tabular}
\caption{Performance comparison of different frequency ranges using ResNet20 structure on CIFAR-10}
\label{tab:kernel}
\end{table}

\begin{figure}[t]
    \centering
    \includegraphics[width=0.80\linewidth]{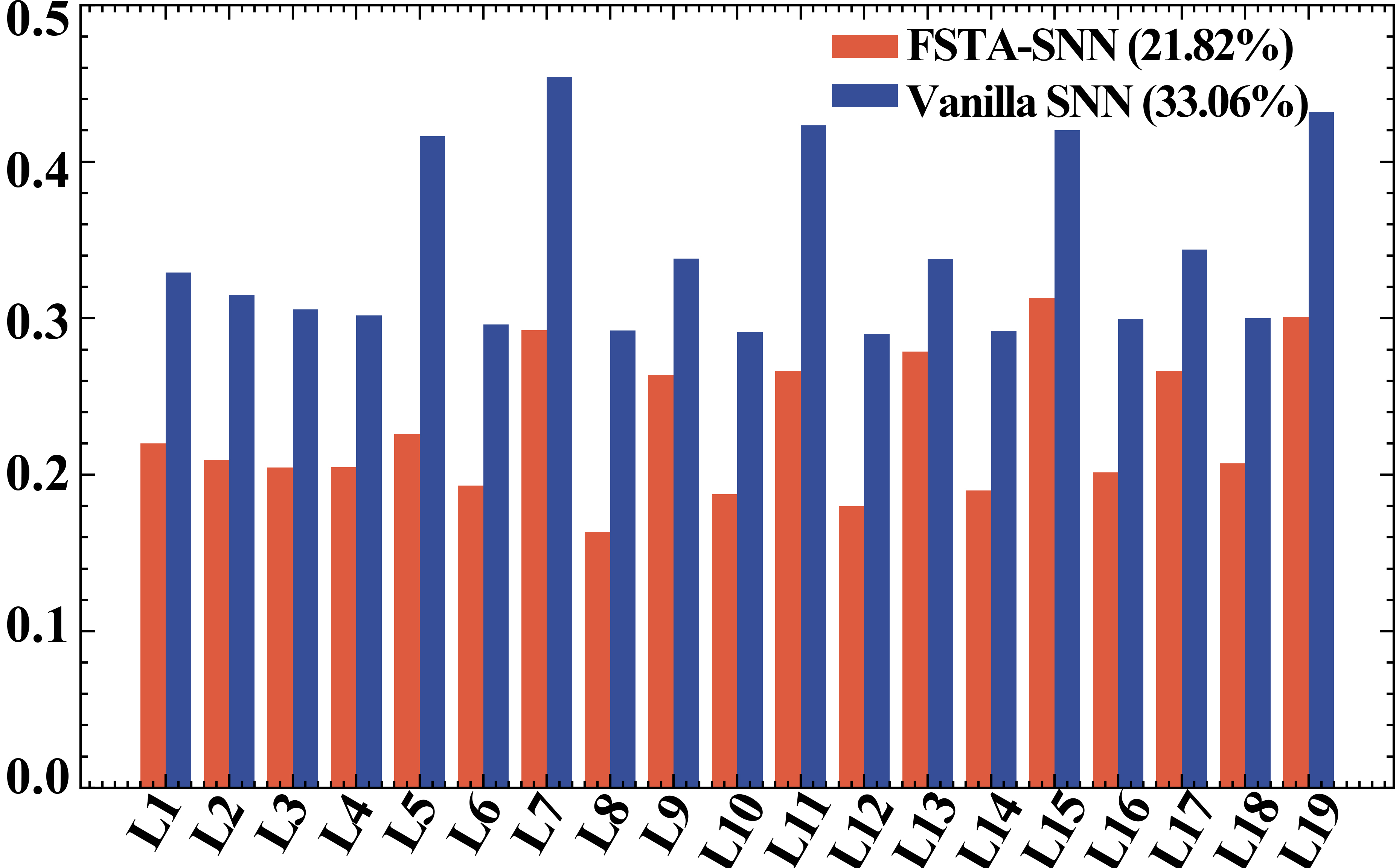}
    \caption{Comparison of spike firing rates at various layers between FSTA-SNN and vanilla SNN. The legend represents the average firing rate of the entire network.}
    \label{fig:albation-spike}
\end{figure}

\subsection{Comparion with SOTA methods}
In this section, we compare our approach with previous SOTA methods.
We report the average top-1 accuracy from these experiments.
Evaluations are first conducted on the static datasets CIFAR-10 and CIFAR-100, with results shown in Tab. \ref{tab:cifar10/100}.
On CIFAR-10, our method improves upon the previous best results by \(0.44\%\) and \(0.05\%\) using ResNet20 and ResNet19, achieving \(94.72\%\) and \(96.52\%\), respectively.
This enhancement is observed even under different timestep conditions.
On CIFAR-100, our method with ResNet19 and ResNet20 at time steps \(2\) and \(4\) yields \(80.42\%\) and \(73.44\%\), respectively, surpassing previous best methods.
Results in Tab. \ref{tab:cifar10/100} clearly demonstrate the superior and efficient performance of our approach.
Additionally, we test our method on the more complex ImageNet dataset, with comparative results shown in Tab. \ref{tab:imagenet}.
Our approach achieves \(68.21\%\) and \(70.23\%\) using the same models, reflecting significant performance improvements over existing methods.
Finally, on the dynamic dataset CIFAR10-DVS, evaluations using ResNet20 achieve \(81.50\%\) and \(82.70\%\) at time steps \(10\) and \(16\), respectively.
This represents a substantial improvement over previous approaches.

\begin{figure}[t]
    \centering
    \includegraphics[width=1.0\linewidth]{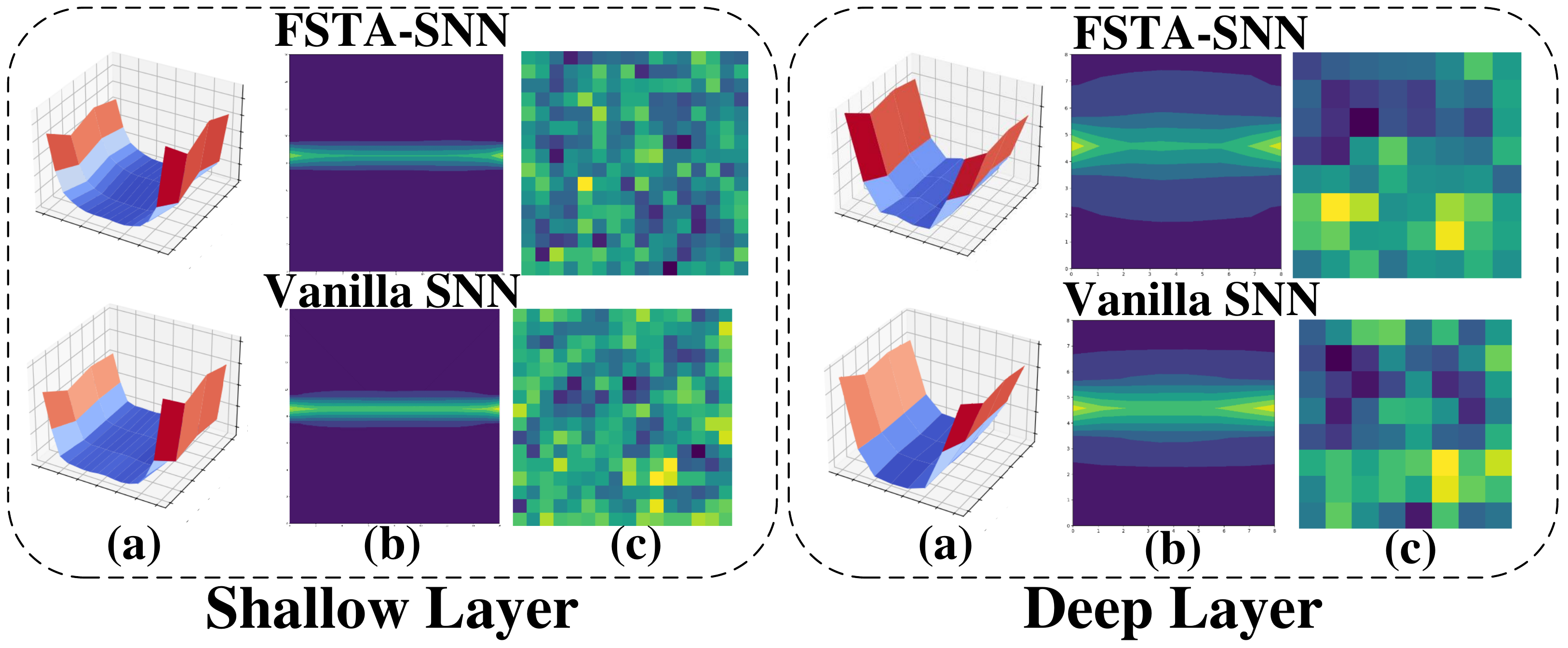}
    \caption{ 
    Visualization of the FSTA-SNN and vanilla SNN analysis.
    (a) Distribution of energy accumulation across frequency bands in the time dimension; (b) Contour plots of the averaged temporal frequency spectrum energy;
    (c) Grad-CAM visualization for the same sample.}
    \label{fig:ablation-analy}
\end{figure}

\subsection{Energy Estimation}
In this section, we evaluate the energy performance of the proposed method.
Fig. \ref{fig:albation-spike} presents a statistical analysis of the spike firing rate in the vanilla SNN compared to our proposed method across various layers.
The results show a substantial reduction in spike firing rate at each layer, with an overall decrease of \(33.99\%\) across the network.
This reduction indicates that our approach effectively eliminates redundant spikes, thereby enhancing the feature representation capability of network.
In Tab. \ref{tab:energy-cost}, we provide detailed data on energy consumption for the model and dataset.
The table indicates that our model introduces only a minimal number of additional parameters while effectively reducing MAC operations and increasing AC computations.
Notably, this performance improvement is achieved without a significant increase in energy consumption.

\subsection{Result Analysis}
We also analyze the spike spectral distribution of the model using DFT with the proposed method.
As shown in Fig. \ref{fig:ablation-analy}, the frequency distribution remains largely consistent with vanilla SNN after incorporating the FSTA module.
However, the magnitudes have significantly decreased. 
This reduction indicates that the proposed method effectively suppresses unnecessary spikes, consistent with the reduced spike firing rate reported in Fig. \ref{fig:albation-spike}.
Fig. \ref{fig:ablation-analy}(b) further illustrates that the introduction of the module narrows the region of shallow spectral energy concentration, reducing redundant spike generation.
In contrast, deeper layers expand the frequency distribution, revealing new spike features that the vanilla SNN could not capture.
The spectral diagrams correspond well with the heat map shown in Fig. \ref{fig:ablation-analy}(c).

\section{Conclusion}
In this paper, we analyze the spike output of SNN from a novel perspective of frequency, revealing the learning preferences of the network.
Based on these findings, we propose a plug-and-play Frequency-Based Spatial-Temporal Attention (FSTA) module.
This module enhances the inherent characteristics of SNNs to improve feature learning capability and reduce redundant spikes.
Experimental results demonstrate that our method significantly enhances model performance across multiple datasets while maintaining low energy consumption.

\section{Acknowledgements}
 This work was supported in part by the National Natural Science Foundation of China (NSFC) under Grant No. 62476035, and 62206037.
 
\bibliography{aaai25}

\end{document}